\PassOptionsToPackage{table}{xcolor}
\documentclass[conference]{IEEEtran}
\IEEEoverridecommandlockouts
\usepackage{cite}
\usepackage{booktabs}
\usepackage{amsmath,amssymb,amsfonts}
\usepackage{algorithm}
\usepackage{algorithmic}
\usepackage{graphicx}
\usepackage{textcomp}
\usepackage{xcolor}
\usepackage{hyperref}
\usepackage{float}
\restylefloat{algorithm}

\usepackage{caption}
\usepackage{subfigure}
\newcommand{\scientific}[2]{#1\,\mathrm{e} {#2}}
\def\BibTeX{{\rm B\kern-.05em{\sc i\kern-.025em b}\kern-.08em
    T\kern-.1667em\lower.7ex\hbox{E}\kern-.125emX}}
\makeatletter
\newcommand{\linebreakand}{%
  \end{@IEEEauthorhalign}
  \hfill\mbox{}\par
  \mbox{}\hfill\begin{@IEEEauthorhalign}
}
\makeatother

\begin{document}

\title{GPU-accelerated Evolutionary Many-objective Optimization Using Tensorized NSGA-III
}

\author{
\IEEEauthorblockN{Hao Li} 
\IEEEauthorblockA{Department of Computer Science\\ and Engineering\\ Southern University of Science\\ and Technology\\  Shenzhen, Guangdong, China \\ li7526a@gmail.com} 
\and 
\IEEEauthorblockN{Zhenyu Liang} 
\IEEEauthorblockA{Department of Computer Science\\ and Engineering\\ Southern University of Science\\ and Technology\\  Shenzhen, Guangdong, China \\ zhenyuliang97@gmail.com} 
\and 
\IEEEauthorblockN{Ran Cheng\textsuperscript{*}\thanks{* Corresponding author}}
\IEEEauthorblockA{Department of Data Science\\ and Artificial Intelligence\\
Department of Computing\\
The Hong Kong Polytechnic University\\ Hong Kong SAR, China \\ ranchengcn@gmail.com}
} 

\maketitle

\begin{abstract}
NSGA-III is one of the most widely adopted algorithms for tackling many-objective optimization problems. 
However, its CPU-based design severely limits scalability and computational efficiency.
To address the limitations, we propose {TensorNSGA-III}, a fully tensorized implementation of NSGA-III that leverages GPU parallelism for large-scale many-objective optimization.
Unlike conventional GPU-accelerated evolutionary algorithms that rely on heuristic approximations to improve efficiency, TensorNSGA-III maintains the exact selection and variation mechanisms of NSGA-III while achieving significant acceleration. 
By reformulating the selection process with tensorized data structures and an optimized caching strategy, our approach effectively eliminates computational bottlenecks inherent in traditional CPU-based and naïve GPU implementations.
Experimental results on widely used numerical benchmarks show that TensorNSGA-III achieves speedups of up to \textbf{$3629\times$} over the CPU version of NSGA-III. Additionally, we validate its effectiveness in multiobjective robotic control tasks, where it discovers diverse and high-quality behavioral solutions. 
Furthermore, we investigate the critical role of large population sizes in many-objective optimization and demonstrate the scalability of TensorNSGA-III in such scenarios. 
The source code is available at \url{https://github.com/EMI-Group/evomo}.
\end{abstract}

\begin{IEEEkeywords}
NSGA-III, Evolutionary Algorithm, GPU Computing, Many-objective Optimization
\end{IEEEkeywords}
\section{Introduction}
In real-world scenarios, it is common to encounter problems that require the simultaneous optimization of multiple conflicting objectives, such as electronic transaction network design~\cite{etn}. These challenges are categorized as multiobjective optimization problems (MOPs) and are commonly formulated as:
\begin{equation}
\label{eq:mop}
\begin{aligned}
\text{Minimize} \quad & F(\mathbf{x}) = \bigl(f_1(\mathbf{x}), f_2(\mathbf{x}), \dots, f_m(\mathbf{x})\bigr), \\
\text{subject to} \quad & \mathbf{x} \in \Omega,
\end{aligned}
\end{equation}
where $\mathbf{x} = (x_1,x_2,\dots,x_d)^{T}$ lies in a $d$-dimensional decision space $\Omega \subseteq \mathbb{R}^d$, $F(\mathbf{x})$ maps $\mathbf{x}$ to an $m$-dimensional objective space $\Theta \subseteq \mathbb{R}^m$. Since the objectives often conflict, no single solution excels in all objectives simultaneously. Instead, decision-makers seek a set of trade-off solutions that balance the competing objectives. This collection of optimal trade-off solutions is known as the Pareto set (PS), and its representation in the objective space forms the Pareto front (PF).

Various approaches have been developed to approximate the PF of MOPs, including scalarization-based methods, gradient-based approaches~\cite{mgda}, and evolution-based approaches known as multiobjective evolutionary algorithms (MOEAs)~\cite{moeas}. MOEAs have garnered significant attention due to their population-based search mechanisms and their ability to perform black-box optimization. Classic MOEAs, such as NSGA-II~\cite{nsga2}, have demonstrated outstanding performance for problems with up to three objectives by utilizing Pareto dominance for solution selection. However, when extending to problems with four or more objectives, termed many-objective optimization problems (MaOPs), traditional methods encounter substantial challenges.

The primary challenges in MaOPs include dominance resistance~\cite{dr} and the exponential increase in required population size. As the number of objectives increases, the likelihood of one solution dominating another decreases sharply, resulting in a proliferation of non-dominated solutions and insufficient selection pressure based on Pareto dominance. Additionally, the PF in an \( m \)-objective problem can be viewed as an \( (m-1) \)-dimensional manifold, necessitating exponentially larger populations to effectively capture its geometry~\cite{MAOP_ACTION}. Traditional population sizes are often inadequate for handling the complexity of MaOPs. Furthermore, visualizing and interpreting solutions in high-dimensional objective spaces pose inherent difficulties~\cite{visionMOP}.

Over the past decades, numerous many-objective evolutionary algorithms (MaOEAs)~\cite{maoeasurvey} have been developed to address these challenges. Among them, NSGA-III~\cite{nsga3} has emerged as one of the most widely used MaOEAs for solving MaOPs and serves as a benchmark for comparing other MaOEAs~\cite{nsga3survey}. NSGA-III enhances the capabilities of NSGA-II by introducing reference points to maintain diversity across the PF. However, like other evolutionary algorithms (EAs), NSGA-III was originally designed for CPU architectures, limiting its computational speed and the population size it can efficiently handle. 

In recent years, the advancements in GPU computing power~\cite{gpgpu} and high-level GPU-acceleration frameworks (e.g., PyTorch~\cite{pytorch}, JAX~\cite{jax}) have revitalized interest in GPU-accelerated evolutionary algorithms~\cite{gpgpu}. These efforts have demonstrated that GPUs can process large populations in parallel, achieving significant speedups over traditional CPU implementations. However, many existing GPU-accelerated evolutionary algorithms rely on hand-written CUDA kernels or only partially optimize code for GPUs, which complicates maintenance and limits scalability for large populations or high-dimensional objective spaces.

Tensorization~\cite{tensorneat, tensorRVEA}, the practice of representing data and operations in tensor (high-dimensional matrix) form, provides a systematic approach to exploit GPU capabilities in EAs. By representing populations and related operations as tensors, it is possible to fully leverage the parallel computation power of GPUs. While prior works~\cite{vectorNSGA2, tensor} indicate that tensorization can significantly accelerate computations and handle larger populations, fully tensorized MaOEAs remain underdeveloped. Moreover, existing tensorized and GPU-accelerated algorithms often approximate the original algorithms due to the challenges in handling the dynamic selection mechanisms required for algorithms like NSGA-III. Additionally, there are few instances of applying tensorized algorithms to real-world problems.

In this paper, we present {TensorNSGA-III}, a fully tensorized implementation of NSGA-III designed to leverage tensor-based computations for many-objective optimization. Our approach addresses two key challenges: (1) reducing the computational overhead associated with dynamic selection in the diversity maintenance stage while preserving the exact selection logic of NSGA-III, and (2) achieving substantial acceleration and scalability to efficiently handle large population sizes in MaOPs.

The primary contributions of this work are as follows:
\begin{itemize}
    \item We introduce TensorNSGA-III, which retains the fundamental selection mechanisms of NSGA-III while achieving superior computational efficiency through full tensorization. Benchmark evaluations demonstrate that TensorNSGA-III achieves speedups of up to \textbf{$3629\times$} compared to the CPU version of NSGA-III.
    \item We conduct a systematic analysis of the interplay between population size and generation count under fixed computational budgets in both continuous and combinatorial optimization settings. Our findings emphasize the critical role of large populations in effectively solving MaOPs.
    \item We validate the effectiveness of TensorNSGA-III in multiobjective robotic control tasks through neuroevolution-based optimization, demonstrating its capability to generate diverse and high-quality behavioral solutions in real-world applications.
\end{itemize}

\section{Background}

\subsection{Many-objective Optimization}
Many-objective optimization problems (MaOPs) typically refer to MOPs with more than three objectives~\cite{MaMOPSurvey}. As the objective dimensionality increases, the selection pressure exerted by Pareto dominance on the population decreases sharply. Consequently, the performance of traditional Pareto-based algorithms may deteriorate due to a preference for extreme solutions, which exhibit poor performance in some objectives while excelling in others. To mitigate this phenomenon, researchers have proposed various methods.

To enhance selection pressure, relaxed dominance-based approaches modify the dominance criteria to expand a solution's dominance space, such as \(\epsilon\)-dominance~\cite{epsdomi}. Indicator-based approaches, on the other hand, utilize performance indicators as fitness functions instead of relying solely on Pareto dominance. By employing different indicators, researchers aim to balance convergence and diversity, achieving a more balanced optimization process, with the hypervolume indicator being a prominent example. Decomposition-based methods aggregate multiple objectives into scalar functions using a set of weighting vectors. NSGA-III is classified as a decomposition-based approach due to its use of reference points, which guide the search direction along these predefined vectors. By improving the diversity management mechanism, NSGA-III compensates for the reduced selection pressure associated with Pareto dominance in MaOPs.

Besides modifying search strategies, increasing the population size can enhance the performance of MaOEAs. Larger populations promote exploration and prevent premature convergence, albeit at the expense of higher computational costs. Theoretical studies on single-objective EAs~\cite{popstudy2} suggest that excessively large populations may have adverse effects under specific conditions. However, these insights are not directly transferable to MOEAs, as simplified models often omit essential mechanisms like diversity maintenance. Experimental research~\cite{positivepop3} in MaOPs indicates that larger populations are generally preferable for effectively capturing high-dimensional Pareto fronts and minimizing the loss of globally competitive solutions due to crowding. Moreover, as computational costs escalate with population size and CPU performance becomes a bottleneck, traditional studies are typically limited to hundreds of individuals. Consequently, GPU acceleration is indispensable for managing populations of thousands or even tens of thousands of individuals.

\subsection{GPU-accelerated MOEAs}
The rapid development of general-purpose GPU computing since 2006 has revolutionized the field of deep learning, and has shown significant potential for accelerating evolutionary algorithms. 
Early implementations of GPU-based MOEAs~\cite{earlyEA} primarily relied on manually optimized CUDA kernels or specialized frameworks, which were often hardware-specific and challenging to extend.

Subsequent research~\cite{gasrea, gpundsort, cudaalgos} focused on decomposing MOEAs into multiple computational tasks that could be executed in parallel across GPU threads. While this approach improved efficiency, it also introduced challenges related to thread divergence, memory access patterns, and load balancing. For instance, Aguilar-Rivera~\cite{vectorNSGA2} developed a GPU-accelerated NSGA-II implementation using an approximate non-dominated sorting method, where each thread processes a subset of the population before merging the results. Additionally, parallelization models such as master-slave~\cite{masterslave} and cellular models~\cite{mocell} have been explored in CPU-based evolutionary algorithms but often require substantial modifications to fully leverage GPU parallelism~\cite{GPUsurvey}.

More recently, a growing body of research~\cite{tensor, matrixEC, evomo} has explored tensor-based approaches for MOEAs, representing populations, variation operators, and selection mechanisms as tensorized computations. Particularly, Liang \textit{et al.}~\cite{tensorRVEA} introduced a fully tensorized RVEA~\cite{rvea}, achieving substantial acceleration on both CPU and GPU architectures when tackling MaOPs. Furthermore, high-level evolutionary computation frameworks such as EvoX~\cite{evox} have emerged to simplify GPU-based evolutionary algorithms by abstracting low-level CUDA operations, thereby enabling efficient parallel execution without requiring specialized GPU programming expertise. 

Tensor-based methods naturally align with modern GPU architectures by leveraging vectorized operations and large-scale parallelism, eliminating the need for low-level memory management. 
However, apart from the very recent work in \textit{et al.}~\cite{tensorRVEA}, this research direction remains largely unexplored.
Additional details on tensor-based EAs are provided in Appendix A.

\subsection{Overview of NSGA-III}
By introducing reference points, NSGA-III extends NSGA-II to better handle MaOPs. Its principal innovation lies in the diversity preservation mechanism during the selection process.

Starting from an initial population \( P \) of size \( n \), NSGA-III iterates through offspring generation and selection stages. At the \( t \)-th generation, offspring generation uses a mating pool (often formed via tournament selection) and variation operators like simulated binary crossover (SBX) and polynomial mutation, producing an offspring population \( O_t \) of size \( n \).

Next, NSGA-III merges the parent population \( P_t \) and the offspring population \( O_t \) into \( R_t = P_t \cup O_t \) of size \(2n\). Similar to NSGA-II, non-dominated sorting is applied to \( R_t \), yielding several fronts \( F_1, F_2, \dots \) ranked by non-domination levels. Individuals within the same front do not dominate each other, and lower fronts indicate fewer domination relationships. The lowest front corresponds to the current non-dominated solutions. The next generation population \( S_t \) is filled by adding entire fronts in ascending order of non-domination rank until its size reaches or exceeds \( n \). If \( |S_t| = n \), the algorithm proceeds to the next iteration with \( P_{t+1} = S_t \). In most cases, however, if \( |S_t| < n \) and \( |S_t \cup F_l| > n \), where \( F_l \) is the next front, NSGA-III must select \( k = n - |S_t| \) individuals from \( F_l \).This procedure is known as niche selection.

\begin{figure}[t]
    \centering
    \includegraphics[width=0.6\linewidth]{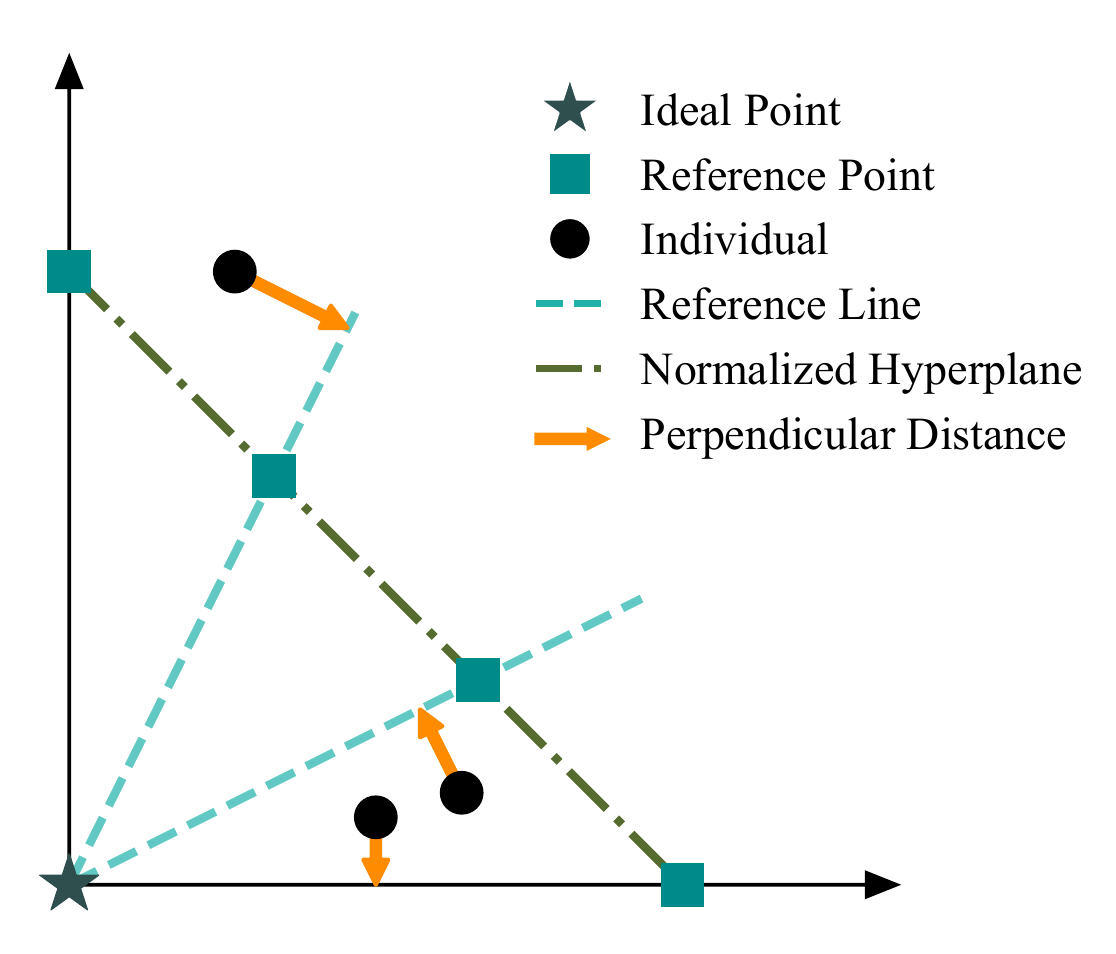}
    \captionsetup{skip=-0.5pt}
    \caption{Association of individuals with reference points in NSGA-III based on perpendicular distance.
    }
    \label{fig:nsga3}
\end{figure}

Unlike NSGA-II, which uses crowding distance to select from \( F_l \), NSGA-III replaces crowding distance with a set of predefined reference points to improve performance in MaOPs. In the absence of prior knowledge, these reference points are uniformly distributed on a normalized hyperplane with number \(w\) which close to \(n\), representing an \((m-1)\)-dimensional hyperplane in an \( m \)-dimensional objective space. Each individual is associated with the reference point to which it has the minimum perpendicular distance (as illustrated in Figure~\ref{fig:nsga3}).

\renewcommand{\algorithmicrequire}{\textbf{Input:}}
\renewcommand{\algorithmicensure}{\textbf{Output:}}
\begin{algorithm}[ht]
\caption{NSGA-III Framework}
\label{alg:nsga3}
\begin{algorithmic}[1]
\REQUIRE Population \( P_t \) with size \( n \), reference point set \( Z \);
\ENSURE Next population \( P_{t+1} \);
\STATE \( Q_t \gets \text{Crossover+Mutation}(P_t) \);
\STATE \( R_t \gets P_t \cup Q_t \);
\STATE \( (F_{s},F_l, F_{drop}) \gets \text{Non-dominated-sort}(R_t) \);
\STATE \( S \gets F_{s} \);
\STATE Associate \( S \text{ and } F_{l} \) with \( Z \) using normalized objectives;
\STATE Get niche count $\rho, \rho' \text{ corresponding } S, F_{l} $;
\WHILE{\( |S| < n\)}
    \STATE Randomly select $v \in Z$ with minimum niche count in \( \rho \);
    \IF{\(\rho'[v] = 0 \)}
        \STATE \( Z \gets Z \setminus \{v\} \);
    \ELSE
        \IF{\(\rho[v] = 0 \)}
            \STATE Select the nearest individual $t \in F_l$ and associated to $v$; 
        \ELSE
            \STATE Randomly select a individual $t \in F_l$ and associated to $v$; 
        \ENDIF
        \STATE Update $F_l, S, \rho, \rho' \text{ by } t$;
    \ENDIF 
\ENDWHILE
\STATE \( P_{t+1} \gets S_t \).
\end{algorithmic}
\end{algorithm}

After associating individuals with reference points, the selection proceeds by iteratively choosing individuals based on reference point occupancy. Within a while-loop, the reference point \( r \) with the fewest associated individuals is selected. If \( r \) has no associated individuals in \( S_t \) but does have individuals in \( F_l \), then the individual in \( F_l \) with the minimum perpendicular distance to \( r \) is chosen and added to \( S_t \). If there are already individuals in \( S_t \) associated with \( r \), a randomly chosen individual from those in \( F_l \) that are associated with \( r \) is added to \( S_t \). This process continues until \( |S_t| = n \). Finally, the next population is set as \( P_{t+1} = S_t \). Algorithm~\ref{alg:nsga3} summarizes the procedure for generation \( t \) of NSGA-III.

\section{Tensorization of NSGA-III}
Tensorization casts all data as tensors, effectively high-dimensional arrays. In a tensor, data is densely represented, with extra information embedded within its dimensions. On GPUs, tensors are efficiently processed by specialized hardware (tensor cores), offering substantial parallel processing capabilities supported by vectorized mathematical and logical operations. In EAs, populations can be naturally represented as tensors, allowing operations to be applied simultaneously to multiple individuals.

In the proposed TensorNSGA-III, tensorization is comprehensively implemented through the tensorized data structures and operators. While many basic tensor operations are well-known in GPU frameworks, fully applying them to MaOEAs involves careful handling of dynamic selection mechanisms. TensorNSGA-III is implemented using EvoX~\cite{evox}, ensuring robust GPU acceleration, simplified memory management and optimization in complex optimization scenarios.

\subsection{Tensorized Data Structures and Basic Operators}
In TensorNSGA-III, individuals and populations are represented using tensor structures that are inherently suited for GPU computation. Specifically, each individual \( \mathbf{x} = [x_1, x_2, \dots, x_d]^T \) in a population of size \( n \) is represented as: \( \mathbf{P} = [\mathbf{x}_1, \mathbf{x}_2, \cdots, \mathbf{x}_n]^T\)

Similarly, the objective values for each individual and the entire population are represented as tensors with shapes \( n \times m \) and \( w \times m \), respectively, where  \( n \) is the population size, \( m \) is the number of objectives and \( w \) is the number of reference points in NSGA-III.

Variation operators such as SBX and polynomial mutation are re-implemented using tensor operations, allowing simultaneous processing of many individuals, as introduced in other work~\cite{tensorRVEA}. Basic tensor operations, including masking, element-wise multiplication (Hadamard product, denoted by \( \circ \)), and the Heaviside step function (denoted by \( H \)), are employed to efficiently implement selection and other algorithmic steps on GPU hardware. The use of $\textit{NaN}$ (Not A Number) as placeholders maintains consistent tensor shapes, facilitating parallel operations on the GPU.

\subsection{Tensorized Selection}
The core innovation of TensorNSGA-III lies in the tensorized selection process, which aims to reduce the while-loop cost presented in the original NSGA-III's selection mechanism. This is achieved by introducing a cache tensor that precomputes and stores the association information between individuals and reference points, as well as the selection order, thereby simplifying the dynamic selection process while maintaining its integrity. 

During the offspring generation stage, TensorNSGA-III mirrors the original algorithm's structure but replaces scalar operations with tensorized counterparts. At the beginning of the selection stage, the merged population \( \mathbf{R_t} = \mathbf{P_t} \cup \mathbf{O_t} \) with shape \( 2n \times d \), and the predefined reference points \( \mathbf{Z} \) with shape \( w \times m \) are shuffled to prepare for random selection during niche selection. The order of objectives tensor \( \mathbf{F_t} \) with shape \( 2n \times m \) remains consistent with the population tensor \( \mathbf{R_t} \). After non dominated sorting, the population is divided into two parts, \( \mathbf{\alpha} \) and \( \mathbf{\beta} \). \( \mathbf{\alpha} \) represents the selected population with ranks smaller than \( l \), and  represents the selecting population with rank \( l \). All rank information record in a tensor \( \mathbf{r}\) with shape \(2n\), the order in \(\mathbf{r}\) corresponding the individuals order in \(\mathbf{R_t}\). The niche selection is outlined in Algorithm~\ref{alg:tensornsga3} corresponding to the next illustration. 

\renewcommand{\algorithmicrequire}{\textbf{Input:}}
\renewcommand{\algorithmicensure}{\textbf{Output:}}
\begin{algorithm}[H]
\caption{Niche Selection Procedure in TensorNSGA-III}
\label{alg:tensornsga3}
\begin{algorithmic}[1]
\REQUIRE Combined population tensor \( \mathbf{R_t} \), objective tensor \( \mathbf{F_t} \), reference point tensor \( \mathbf{Z} \), rank tensor \(\mathbf{r}\) and selecting rank \( l \).
\ENSURE Next population \( \mathbf{P_{t+1}} \)

\STATE Normalize objective tensor: \( \mathbf{F_t} \gets \text{Normalize}(\mathbf{F_t}) \);
\STATE Get perpendicular distances tensor \(\mathbf{r}\) by Equation~\ref{eq:dist};
\STATE Determine nearest reference points: \( \mathbf{\pi} \gets \arg \min (\mathbf{D}, \text{axis}=1) \);
\STATE Find minimum distances: \( \mathbf{d} \gets \min(\mathbf{D}, \text{axis}=1) \);
\STATE Get count tensor of \(\alpha\): \( \mathbf{\rho} \gets \text{count}(\mathbf{r}<l, \mathbf{\pi}) \);
\STATE Get count tensor of \(\beta\): \( \mathbf{\rho'} \gets \text{count}(\mathbf{r}=l, \mathbf{\pi}) \);
\STATE Update count tensor of \(\alpha\): \( \mathbf{\rho} \gets \mathbf{\rho} + \infty \circ (\mathbf{\rho'} = 0) \);
\STATE Select individuals: \( \mathbf{x} \gets \arg \min(\mathbf{d} \circ (\mathbf{\pi} = j)) \text{ for } j \in (\mathbf{\rho} = 0) \);
\STATE Update rank tensor: \( \mathbf{r} \gets \mathbf{r} - H(\mathbf{r}[\mathbf{x}]) \);
\STATE Update selecting population counts: \( \mathbf{\rho'} \gets \mathbf{\rho'} - (\mathbf{\rho} = 0) \);
\STATE Update reference point counts: \( \mathbf{\rho} \gets \mathbf{\rho} + (\mathbf{\rho} = 0) \);
\STATE Update count tensor: \( \mathbf{\rho} \gets \mathbf{\rho} + \infty \cdot (\mathbf{\rho'} = 0) \);
\STATE Organize cache tensor \( \mathbf{Q} \): each row contains individuals in \( \mathbf{\beta} \) associated with a reference point;
\STATE Initialize selection state tensor: \( \mathbf{s} \gets [1,1,\cdots,1]\);
\WHILE{\( \sum H(l - \mathbf{R}) < K \)}
    \STATE Get mark tensor \( \mathbf{u} \gets \mathbf{\rho} = \min(\mathbf{\rho}) \);
    \STATE Get temporary indicator tensor \( \mathbf{t} \gets \mathbf{u} \circ \mathbf{s} \);
    \STATE Select individuals: \( \mathbf{x} \gets \mathbf{Q}[\mathbf{t}]\);
    \STATE Update rank tensor: \( \mathbf{r} \gets \mathbf{r} - H(\mathbf{r}[\mathbf{x}]) \);
    \STATE Update indicator tensor: \( \mathbf{s} \gets \mathbf{s} + H(\mathbf{s}[\mathbf{x}]) \);
    \STATE Update count tensor of \(\alpha\): \( \mathbf{\rho'} \gets \mathbf{\rho'} - \mathbf{u} \);
    \STATE Update count tensor of \(\beta\): \( \mathbf{\rho} \gets \mathbf{\rho} + \mathbf{u} \);
    \STATE Update count tensor of \(\alpha\): \( \mathbf{\rho} \gets \mathbf{\rho} + \infty \cdot (\mathbf{\rho'} = 0) \);
\ENDWHILE
\STATE Truncate last selection to ensure \( \sum H(\mathbf{R} < l) = n \);
\STATE Get next population: \( \mathbf{P_{t+1}} \gets \mathbf{P_t} \circ H(\mathbf{R} < l) \);
\end{algorithmic}
\end{algorithm}

\paragraph{Selection Preparing (Procedure Lines 1-7)} Before niche selection, the distance between all individuals and all reference points is calculated using the following equation:
\begin{equation}
\label{eq:dist}
\mathbf{D} \gets ||\mathbf{F_t}|| \cdot \sqrt{1 - \frac{\mathbf{F_t} \cdot \mathbf{Z}}{||\mathbf{F_t}|| \cdot ||\mathbf{Z}||}},
\end{equation}
where \( \mathbf{F_t} \) is the normalized objective tensor. In \( \mathbf{D} \) with shape \( 2n \times w \), each row corresponds to an individual in \( \mathbf{R_t} \), and the \( i \)-th value represents the perpendicular distance to the \( i \)-th reference point in \( \mathbf{Z} \). By selecting the minimum value of each row and recording the position information, the association information is collected in the distance tensor \( \mathbf{d} \) and the associated reference tensor \( \mathbf{\pi} \), both with shape \( 2n \).

Next, the number of associations for each reference point is counted in tensors \( \mathbf{\rho} \) and \( \mathbf{\rho'} \) for populations \( \mathbf{\alpha} \) and \( \mathbf{\beta} \), respectively. The reference points counter is zero in \( \mathbf{\mathbf{\rho'}} \) will be set as \(\infty\) in \(\mathbf{\rho}\) to avoid selection.

\paragraph{Nearest Selection (Procedure Lines 8-12)} All reference points without any associated individuals in \( \mathbf{\alpha} \) are processed. This means that the \( i \)-th reference point \(\in \mathbf{Z} \) has a count of zero in tensor \( \mathbf{\rho} \) and a count greater than zero in tensor \( \mathbf{\rho'} \) if the \( i \)-th reference point has been selected. For each such reference point, the nearest individual from \( \mathbf{\beta} \) is selected based on the distance tensor \( \mathbf{d} \). Adding the selected individual to \( \mathbf{\alpha} \) involves updating the rank tensor \( \mathbf{r} \) with \( l-1 \) at the corresponding position. Similarly, tensors \( \mathbf{\rho} \) and \( \mathbf{\rho'} \) are updated accordingly.

\paragraph{Random Selection (Procedure Lines 13-25)} A cache tensor \( \mathbf{Q} \) with shape \( w \times n \) is organized such that each row contains individuals in \( \mathbf{\beta} \) associated with a specific reference point. Corresponding to tensor \( \mathbf{Q} \), a one-dimensional tensor \( \mathbf{s} \) with shape \( w \) is initialized with ones, indicating the next selection index for each reference point's row in \( \mathbf{Q} \).

\begin{figure}[htbp]
    \centering
    \includegraphics[width=0.5\linewidth]{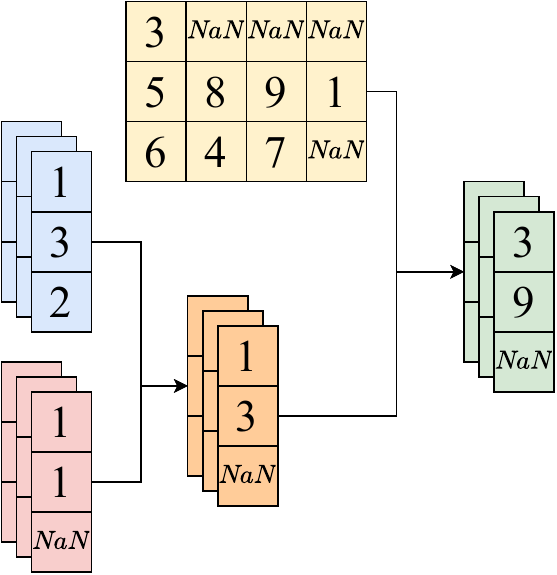}
    \caption{Random Selection in TensorNSGA-III: The indicator tensor (blue) and mark tensor (red) form a temporary indicator (orange), which combines with the cache tensor (yellow) to determine selected individuals (green).}
    \label{fig:tensornsga3}
\end{figure}

In the subsequent selection loop, batches of reference points are processed concurrently using \( \mathbf{Q} \), as illustrated in Figure~\ref{fig:tensornsga3}. Each iteration performs the following steps:
\begin{enumerate}
    \item \textbf{Generate Mark Tensor \( \mathbf{u} \)}: Identify the reference points with the minimum occupancy in population \( \mathbf{\alpha} \). This results in a mark tensor \( \mathbf{u} \) indicating which reference points to select.
    \item \textbf{Combine Mark and Indicator Tensors}: Combine the mark tensor \( \mathbf{u} \) with the selection indicator tensor \( \mathbf{s} \) to obtain the temporary indicator tensor \( \mathbf{t} \) for this iteration.
    \item \textbf{Select Individuals from Cache Tensor \( \mathbf{Q} \)}: Using the temporary indicator tensor \( \mathbf{t} \), select the corresponding individuals from tensor \( \mathbf{Q} \), which indicates their positions in tensor \( \mathbf{r} \).
    \item \textbf{Update Tensors}: Update the rank tensor \( \mathbf{R} \), the selection state tensor \( \mathbf{s} \), and the count tensors \( \mathbf{\rho} \) and \( \mathbf{\rho'} \) based on the selected individuals and selected reference points.
\end{enumerate}
End the loop, until selected individuals' size larger or equal to \(n\). And truncate the last selection to make the selected individuals' size equal to \(n\). The truncation is randomly due to the shuffled reference points. 


\section{Experiments}
This section presents a comprehensive evaluation of the proposed TensorNSGA-III. All experiments were conducted using the EvoX platform~\cite{evox} on a computing system equipped with an AMD EPYC 9654 96-core processor and an NVIDIA\textsuperscript{\textregistered} Tesla V100 GPU.

\begin{table*}[h!]
\centering
\small
\caption{Comparison of average runtime per generation of algorithms on DTLZ3 problem with varying population sizes. Speedup values are compared to NSGA-III on CPU. TensorRVEA are included as supplemental information for reference.}
\label{tab:dtlz3}
\setlength{\tabcolsep}{5pt}
\renewcommand{\arraystretch}{1.2}
\begin{tabular}{l c c c c c}
\hline
\textbf{Population} & \textbf{NSGA-III on GPU} & \textbf{NSGA-III on CPU} & \textbf{TensorNSGA-III} & \textbf{Speedup} & \textbf{TensorRVEA} \\ \hline
50    & $\scientific{4.257}{-2} \pm \scientific{1.25}{-3}$ & $\scientific{4.047}{-3} \pm \scientific{1.43}{-4}$ & \cellcolor{gray!20}{$\scientific{1.120}{-3} \pm \scientific{1.38}{-4}$} & $4$    & $\scientific{2.258}{-3} \pm \scientific{7.32}{-4}$ \\ \hline
100   & $\scientific{1.477}{-1} \pm \scientific{4.87}{-3}$ & $\scientific{9.604}{-3} \pm \scientific{1.45}{-4}$ & \cellcolor{gray!20}{$\scientific{1.135}{-3} \pm \scientific{1.38}{-4}$} & $8$    & $\scientific{3.387}{-3} \pm \scientific{7.97}{-4}$ \\ \hline
200   & $\scientific{6.767}{-1} \pm \scientific{1.78}{-2}$ & $\scientific{3.635}{-2} \pm \scientific{8.79}{-4}$ & \cellcolor{gray!20}{$\scientific{1.121}{-3} \pm \scientific{1.20}{-4}$} & $32$   & $\scientific{5.454}{-3} \pm \scientific{1.22}{-3}$ \\ \hline
400   & $\scientific{2.795}{0} \pm \scientific{7.73}{-2}$ & $\scientific{1.271}{-1} \pm \scientific{1.25}{-3}$ & \cellcolor{gray!20}{$\scientific{1.243}{-3} \pm \scientific{1.18}{-4}$} & $102$  & $\scientific{5.264}{-3} \pm \scientific{1.05}{-3}$ \\ \hline
800   & $\scientific{1.064}{1} \pm \scientific{2.76}{-1}$ & $\scientific{5.000}{-1} \pm \scientific{1.88}{-3}$ & \cellcolor{gray!20}{$\scientific{1.493}{-3} \pm \scientific{1.29}{-4}$} & $335$  & $\scientific{8.003}{-3} \pm \scientific{1.46}{-3}$ \\ \hline
1600  & $\scientific{3.434}{1} \pm \scientific{9.48}{-1}$ & $\scientific{1.595}{0} \pm \scientific{7.52}{-3}$ & \cellcolor{gray!20}{$\scientific{2.063}{-3} \pm \scientific{1.22}{-4}$} & $773$  & $\scientific{8.751}{-3} \pm \scientific{1.60}{-3}$ \\ \hline
3200  & $\scientific{1.609}{2} \pm \scientific{4.84}{0}$ & $\scientific{7.202}{0} \pm \scientific{4.36}{-2}$ & \cellcolor{gray!20}{$\scientific{4.886}{-3} \pm \scientific{1.34}{-4}$} & $1474$ & $\scientific{1.417}{-2} \pm \scientific{2.17}{-3}$ \\ \hline
6400  & $\scientific{6.503}{2} \pm \scientific{1.62}{1}$ & $\scientific{2.990}{1} \pm \scientific{1.64}{-1}$ & \cellcolor{gray!20}{$\scientific{1.575}{-2} \pm \scientific{1.70}{-4}$} & $1898$ & $\scientific{1.824}{-2} \pm \scientific{2.26}{-3}$ \\ \hline
12800 & $\scientific{2.701}{3} \pm \scientific{5.81}{1}$ & $\scientific{2.165}{2} \pm \scientific{1.26}{0}$ & $\scientific{5.966}{-2} \pm \scientific{1.48}{-4}$ & $3629$ & \cellcolor{gray!20}{$\scientific{3.431}{-2} \pm \scientific{3.08}{-3}$} \\ \hline
\end{tabular}
\end{table*}

\subsection{Acceleration Performance}
\label{sec:acceleration}
To validate the acceleration performance of proposed TensorNSGA-III when handling large population sizes on large-scale MaOPs, we compare NSGA-III on CPU, NSGA-III on GPU, and TensorNSGA-III on GPU. Additionally, TensorRVEA on GPU serves as a reference GPU-accelerated algorithm.

For these experiments, we used the DTLZ3~\cite{dtlz} problem with six objectives and 500 decision variables. The population size varied from 50 to 12{,}800, and each algorithm was run for 100 generations. We set the maximum runtime limit for a single experimental configuration to 8~hours and repeated each configuration 31 times, recording the average computation time per generation.

Table~\ref{tab:dtlz3} summarizes the average runtime results per generation. 
The CPU-based NSGA-III struggles with increasing population size. The GPU version performs even worse due to its dynamic selection procedure, which hinders efficient parallelism even with XLA optimization.
In contrast, TensorNSGA-III demonstrates up to $3629\times$ speedup compared to the CPU baseline. 
Comparing TensorNSGA-III with TensorRVEA shows its speed advantage over other tensorized algorithms while validating the cache mechanism's effectiveness.
Before a population size of 12{,}800, TensorNSGA-III with the while-loop consistently runs faster than TensorRVEA without it. This demonstrates that, despite while-loops being unsuitable for GPUs, TensorNSGA-III still achieves high parallel throughput due to the cache mechanism. Notably, at smaller scales, increasing population size minimally affects runtime for both tensorized algorithms. In our experiments, their runtimes stay below 0.6 ms, demonstrating excellent scalability as population size grows.

\begin{figure*}[htbp]

    \centering
     
    \subfigure[DTLZ2]{
        \includegraphics[width=0.28\textwidth]{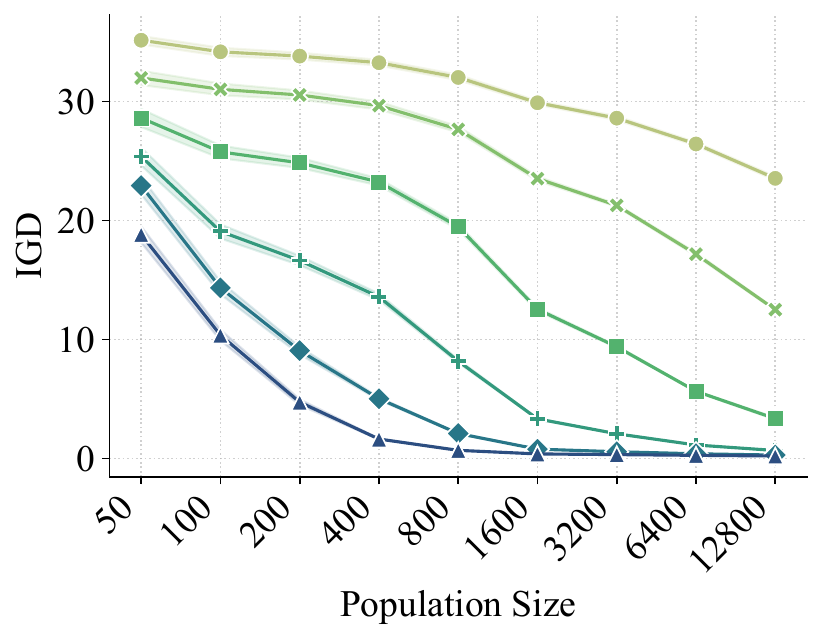}
    }
    \subfigure[DTLZ5]{
        \includegraphics[width=0.28\textwidth]{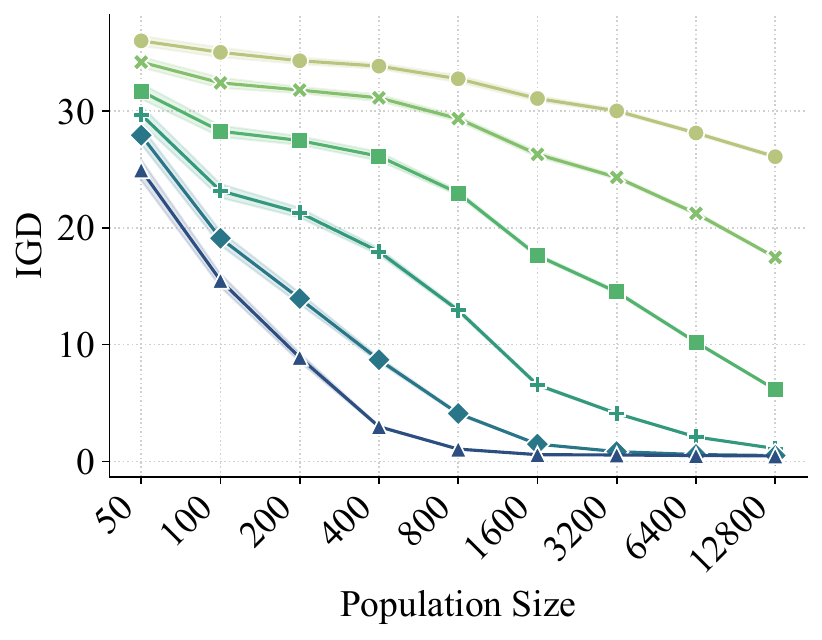}
    }
    \subfigure[DTLZ7]{
        \includegraphics[width=0.28\textwidth]{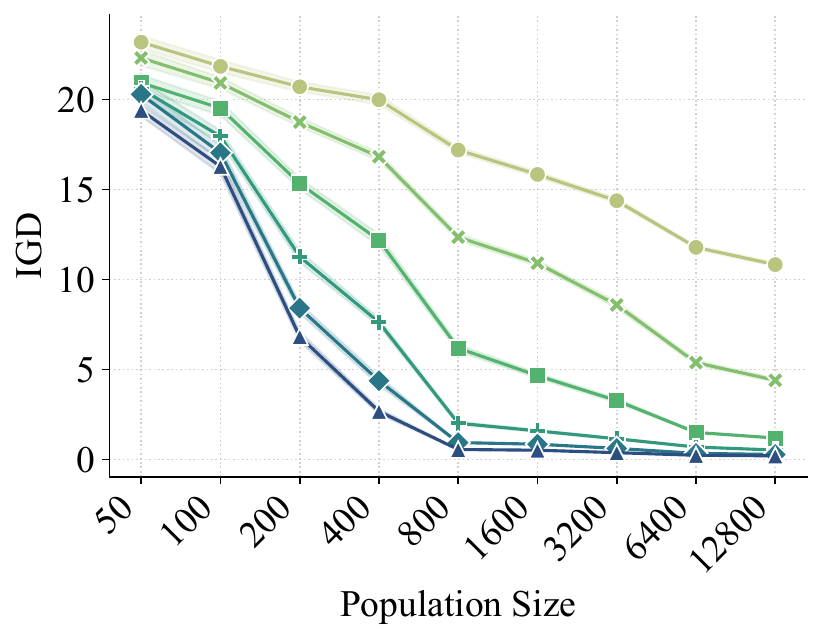}
    }
    \vspace{-1em}
    \subfigure{\includegraphics[width=0.9\textwidth]{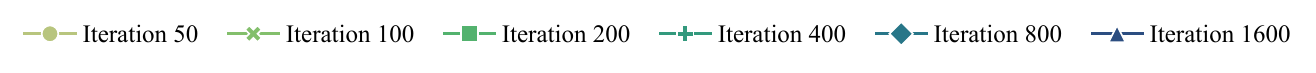}}
    \vspace{-1em}
    \caption{Mean IGD values and 95\% confidence intervals for different population sizes and iterations on DTLZ2, DTLZ5, DTLZ7 using TensorNSGA-III.}
    \label{fig:igdvsgen}
\end{figure*}

\begin{figure}[htbp]

    \centering
      \subfigure[MNKLandscape]{
        \includegraphics[width=0.22\textwidth]{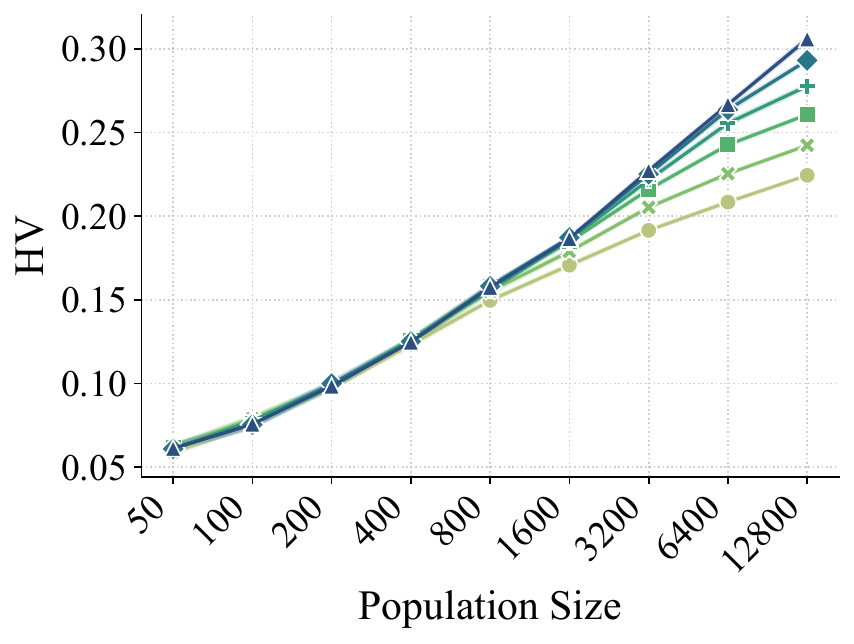}
    }
    \subfigure[MOKnapsack]{
        \includegraphics[width=0.22\textwidth]{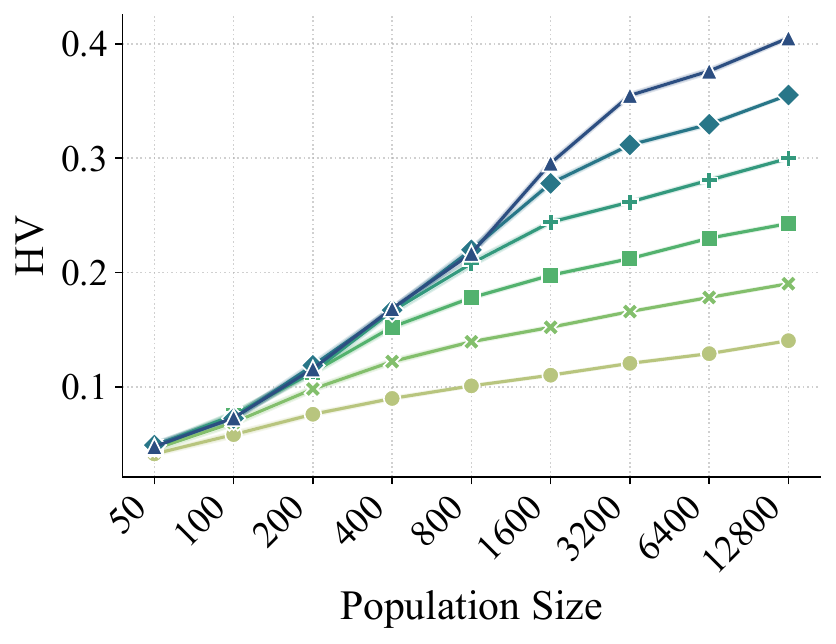}
    }
    \vspace{-1em}
    \subfigure{
        \includegraphics[width=0.48\textwidth]{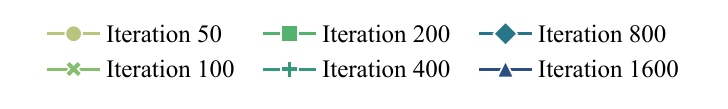}
    }
    \vspace{-1em}
    \caption{Mean HV values and 95\% confidence intervals for different population sizes and iterations on MNKLandscape, MOKnapsack using TensorNSGA-III.}
    \label{fig:hvvsgen}
\end{figure}

\subsection{Impact of Large Populations}
To investigate the impact of large populations on MaOPs, we examined the evolution of IGD and HV under different population sizes and generation counts.
Experiments were conducted with population sizes up to 12{,}800 across various problem types, covering continuous (DTLZ2, DTLZ5, DTLZ7) and discrete (MNKLandscape~\cite{molp}, MOKnapsack~\cite{mokp}) MaOPs.
The DTLZ problems are set with 6 objectives and 500 decision variables.

As shown in Figure~\ref{fig:igdvsgen}, larger population sizes significantly enhance IGD reduction, demonstrating their critical role in achieving better convergence. Increasing the population size from 50 to 800 leads to the most substantial improvement, while further expansion beyond 3200 continues to reduce IGD but with diminishing returns. This trend highlights that larger populations provide better diversity and exploration capability, which is crucial for maintaining a well-distributed PF. The results also show that while higher iteration counts improve convergence, large populations accelerate this process, achieving lower IGD even at earlier iterations. Among the tested problems, DTLZ2 benefits the most from large populations, while DTLZ7 starts with lower IGD, indicating a less complex landscape. The 95\% confidence intervals further suggest that larger populations yield more stable solutions with reduced variance. These findings underscore the advantages of large population sizes in improving solution quality, accelerating convergence, and enhancing stability, making them a key factor in optimizing performance despite increasing computational costs.

\begin{figure*}[hbt]
    \centering
    \subfigure[MoSwimmer]{
        \includegraphics[width=0.28\textwidth]{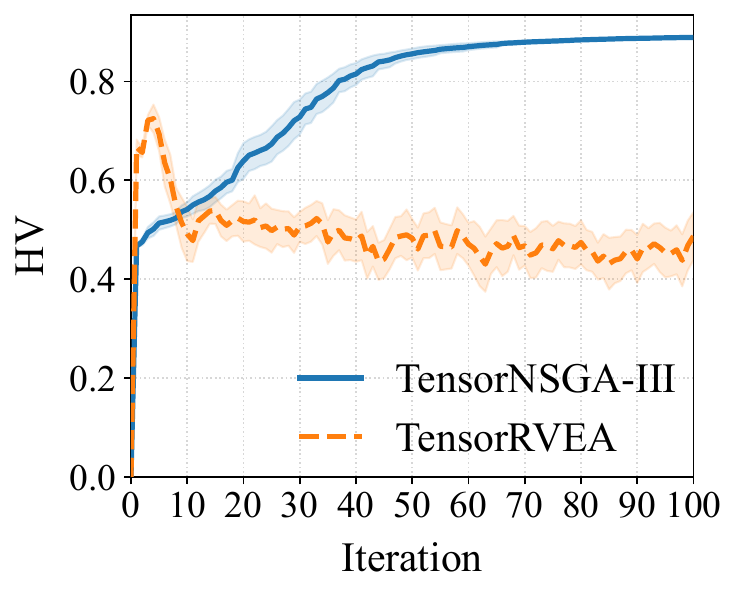}
    }
    \subfigure[MoHalfcheetah]{
        \includegraphics[width=0.28\textwidth]{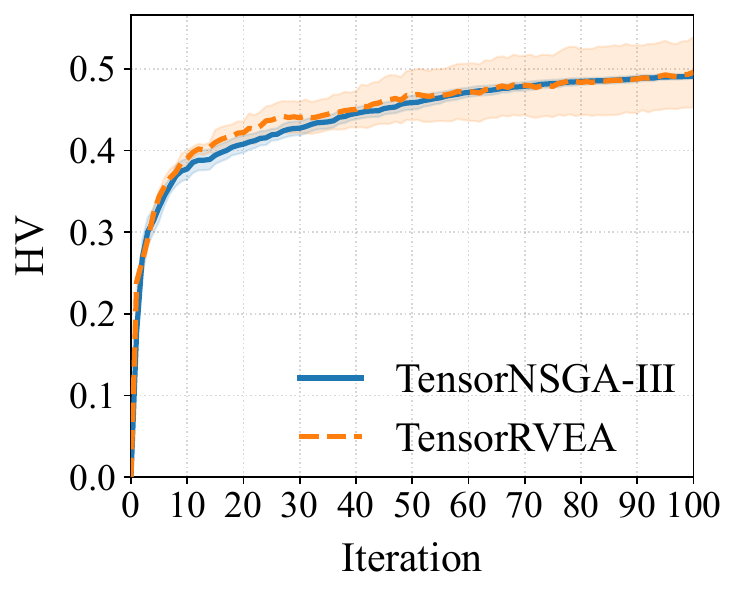}
    }
    \subfigure[MoReacher]{
        \includegraphics[width=0.28\textwidth]{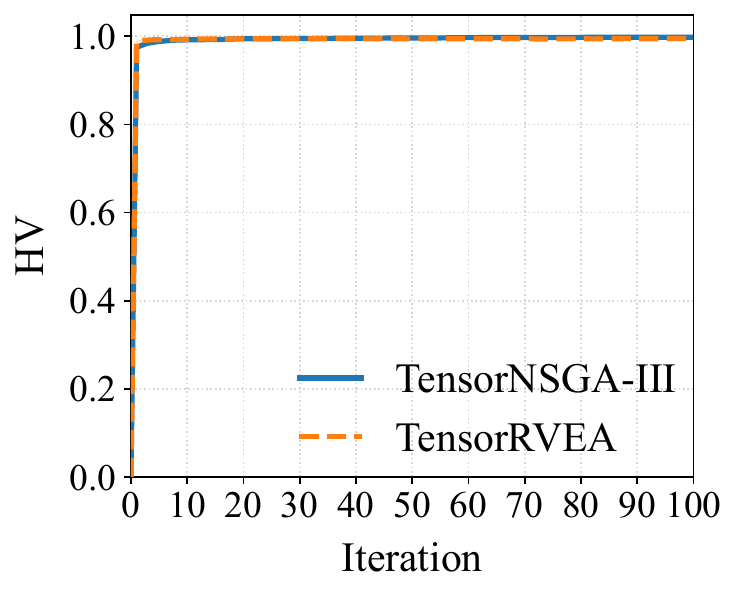}
    }
    \vspace{-1em}
    \caption{Comparison of HV values and 95\% confidence intervals on MoSwimmer, MoHalfcheetah and MoHopper between TensorNSGA-III and basic search.}
    \label{fig:brax1}
\end{figure*}

\begin{figure*}[hbt]
    \vspace{-1em}
    \centering
    \subfigure[MoSwimmer]{
        \includegraphics[width=0.28\textwidth]{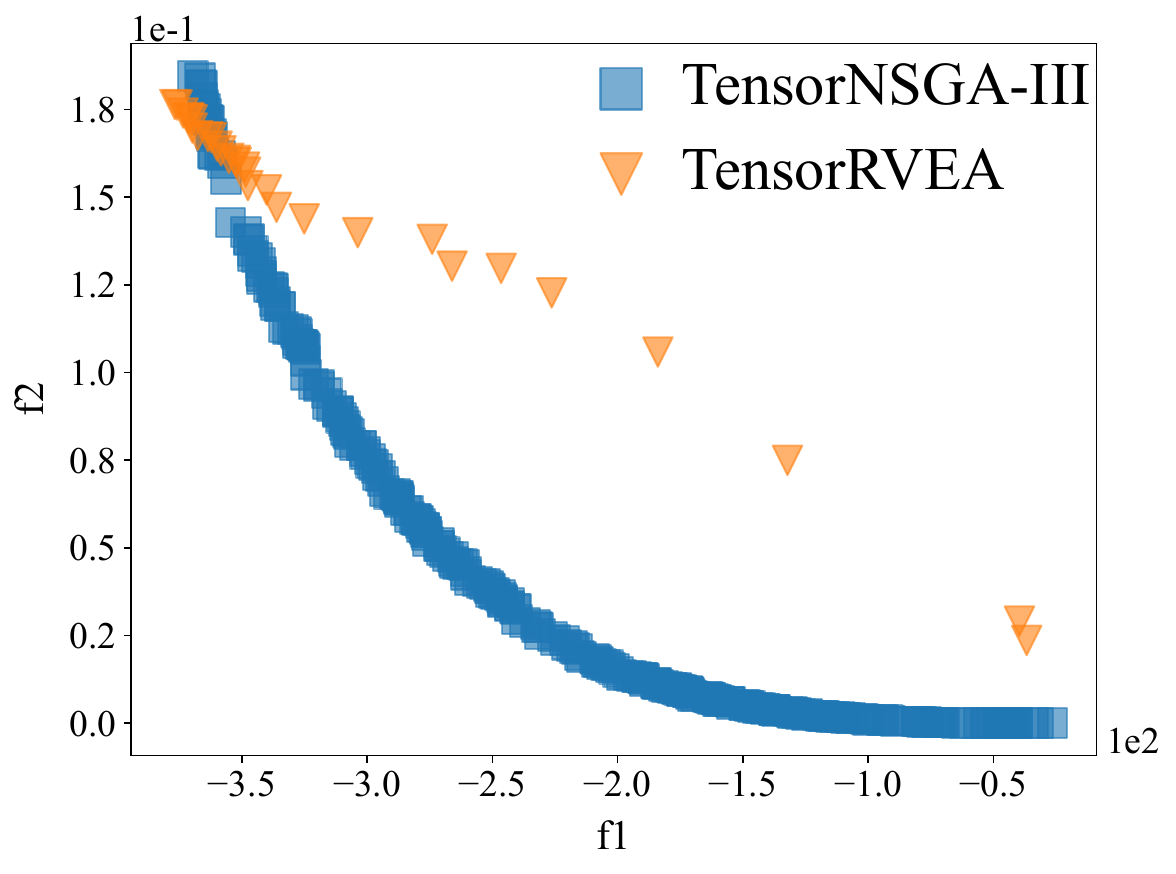}
    }
    \subfigure[MoHalfcheetah]{
        \includegraphics[width=0.28\textwidth]{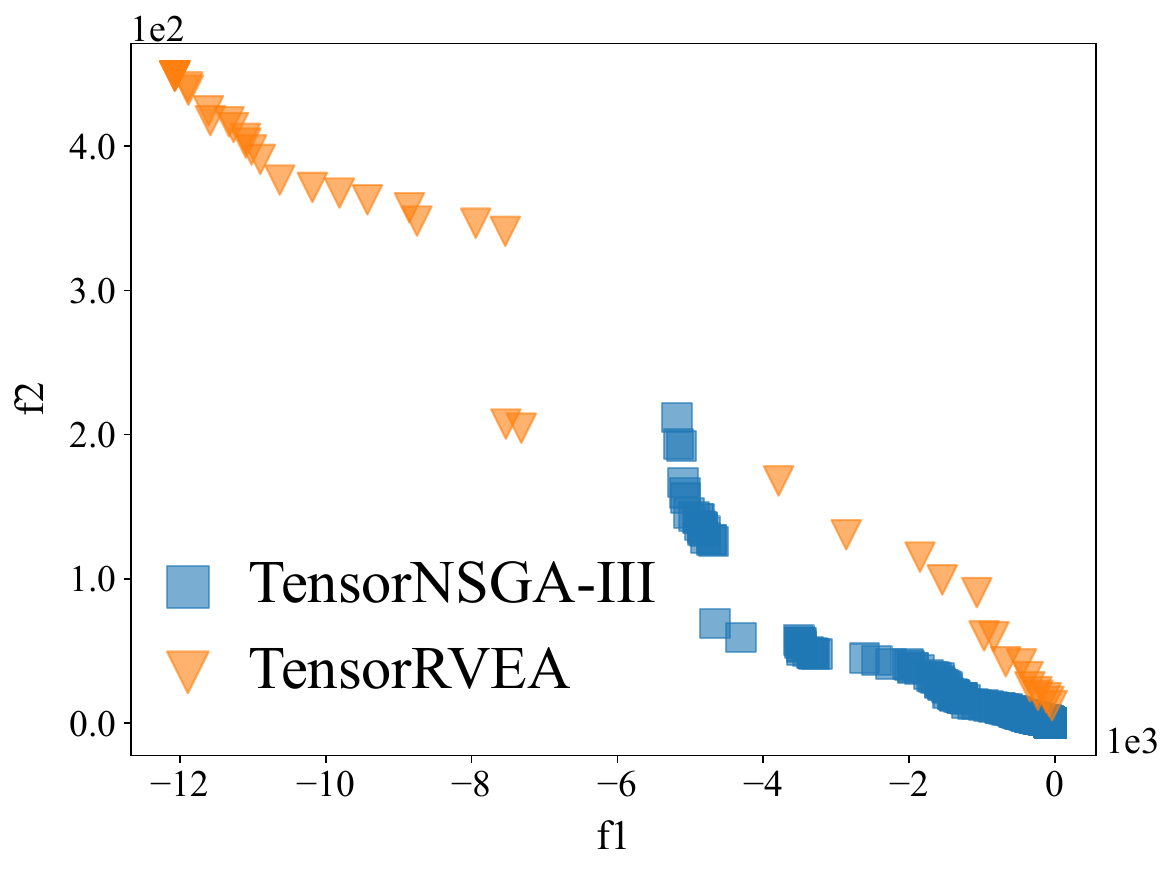}
    }
    \subfigure[MoReacher]{
        \includegraphics[width=0.28\textwidth]{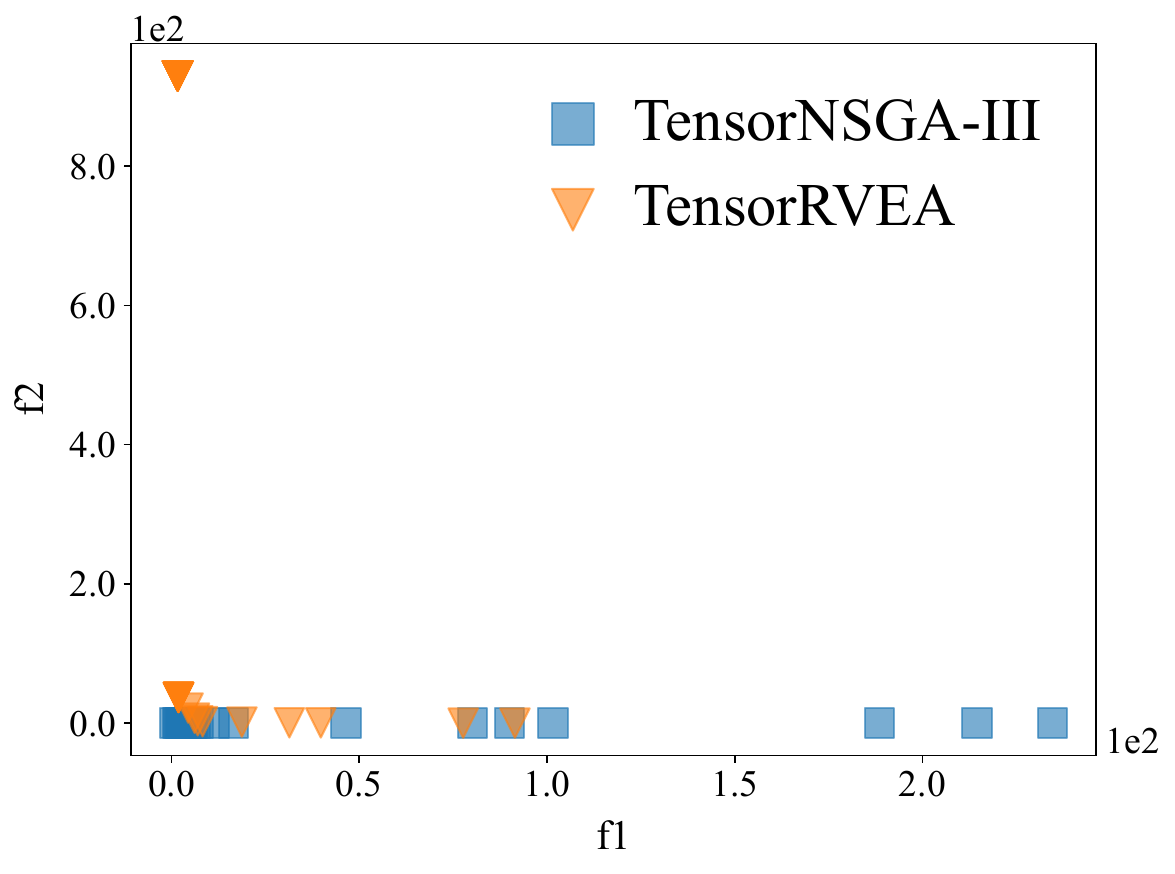}
    }
    \vspace{-1em}
    \caption{Visualization of final solutions on MoSwimmer, MoHalfcheetah and MoReacher between TensorNSGA-III and Basic Search. Lower objective values indicate better performance.}
    \label{fig:brax2}
\end{figure*}

In Figure~\ref{fig:hvvsgen}, the HV trends for MNKLandscape and MOKnapsack are more evident. As expected, when the number of generations is fixed, larger populations generally yield better HV values. In particular, a population size of 12{,}800 achieves the best overall performance across different runs. Compared to the search capability provided by iterations, the population size is more important. For the MNKLandscape problem, before a population size of 800, increasing the number of iterations has little effect on HV improvement. Only with population sizes greater than 800 does increasing the number of iterations become meaningful.

\subsection{Application in Robot Control Tasks}
\label{sec:neuroevolution}
To demonstrate the practical applicability of large-scale GPU-based MaOEAs, we compared TensorNSGA-III with TensorRVEA~\cite{tensorRVEA} on multiobjective robot control tasks using the Brax~\cite{brax} environment. Specifically, we evaluated three multiobjective robot control problems proposed in~\cite{tensorRVEA}, with detailed problem descriptions provided in the appendix.

Inspired by multiobjective neuroevolution, we employed a Multi-Layer Perceptron (MLP) as the policy network, optimizing its weights through population-based search. We set the population size to 1{,}000 and generation number to 100. All algorithms were executed independently 16 times to ensure statistical robustness.

Figure~\ref{fig:brax1} illustrates the progression of Hypervolume (HV) values over generations for each robot control task, while Figure~\ref{fig:brax2} presents the final Pareto Fronts (PFs). In the MoHalfcheetah and MoReacher environments, TensorNSGA-III achieved performance comparable to TensorRVEA in exploring the high-dimensional objective spaces, attaining similar final HV values and demonstrating comparable stability, as evidenced by their confidence intervals. Notably, in the MoSwimmer environment, TensorNSGA-III significantly outperformed TensorRVEA, achieving higher HV values, higher stability and exhibiting greater diversity in the final PF. This indicates that TensorNSGA-III is particularly effective in generating a wide range of control policies that accommodate different task preferences and trade-offs. These results highlight the benefits of utilizing large populations and GPU-accelerated selection mechanisms in complex real-world optimization tasks.

\section{Conclusion}
In this work, we introduced TensorNSGA-III, a tensorized implementation of NSGA-III designed to address the challenges of many-objective optimization with large population sizes. Our findings reveal that naive attempts to accelerate NSGA-III on GPUs without tensorization can yield suboptimal performance. This highlights that simply porting a CPU-based algorithm to the GPU without rethinking its design may be ineffective.

Our approach demonstrates the advantages of tensorization and GPU acceleration. TensorNSGA-III leverages a fully tensorized design to achieve remarkable computational speed and performance improvements without any approximations. The robustness and efficiency of TensorNSGA-III were validated across a diverse range of optimization tasks, including traditional numerical benchmarks, combinatorial optimization problems, and high-dimensional neuroevolution challenges in robot control tasks. The experimental results underscore the significant speedups and enhanced performance of TensorNSGA-III, attributable to its tensorized design and innovative cache mechanism. Moreover, the ability to handle large population sizes suggests that these methods have the potential to tackle even more complex optimization challenges.

\bibliographystyle{IEEEtran}
\bibliography{references}

\begin{thebibliography}{10}
\providecommand{\url}[1]{#1}
\csname url@samestyle\endcsname
\providecommand{\newblock}{\relax}
\providecommand{\bibinfo}[2]{#2}
\providecommand{\BIBentrySTDinterwordspacing}{\spaceskip=0pt\relax}
\providecommand{\BIBentryALTinterwordstretchfactor}{4}
\providecommand{\BIBentryALTinterwordspacing}{\spaceskip=\fontdimen2\font plus
\BIBentryALTinterwordstretchfactor\fontdimen3\font minus \fontdimen4\font\relax}
\providecommand{\BIBforeignlanguage}[2]{{%
\expandafter\ifx\csname l@#1\endcsname\relax
\typeout{** WARNING: IEEEtran.bst: No hyphenation pattern has been}%
\typeout{** loaded for the language `#1'. Using the pattern for}%
\typeout{** the default language instead.}%
\else
\language=\csname l@#1\endcsname
\fi
#2}}
\providecommand{\BIBdecl}{\relax}
\BIBdecl

\bibitem{etn}
C.~Yeh, ``An improved {NSGA2} to solve a bi-objective optimization problem of multi-state electronic transaction network,'' \emph{Reliab. Eng. Syst. Saf.}, vol. 191, 2019.

\bibitem{mgda}
J.-A. Désidéri, ``Multiple-gradient descent algorithm ({MGDA}) for multiobjective optimization,'' \emph{Comptes Rendus Mathematique}, vol. 350, no.~5, pp. 313--318, 2012.

\bibitem{moeas}
A.~Zhou, B.~Qu, H.~Li, S.~Zhao, P.~N. Suganthan, and Q.~Zhang, ``Multiobjective evolutionary algorithms: {A} survey of the state of the art,'' \emph{Swarm Evol. Comput.}, vol.~1, no.~1, pp. 32--49, 2011.

\bibitem{nsga2}
K.~Deb, S.~Agrawal, A.~Pratap, and T.~Meyarivan, ``A fast and elitist multiobjective genetic algorithm: {NSGA-II},'' \emph{{IEEE} Trans. Evol. Comput.}, vol.~6, no.~2, pp. 182--197, 2002.

\bibitem{dr}
C.~M. Fonseca and P.~J. Fleming, ``Multiobjective optimization and multiple constraint handling with evolutionary algorithms. i. a unified formulation,'' \emph{IEEE Transactions on Systems, Man, and Cybernetics-Part A: Systems and Humans}, vol.~28, no.~1, pp. 26--37, 1998.

\bibitem{MAOP_ACTION}
H.~Ishibuchi, N.~Akedo, and Y.~Nojima, ``Behavior of multiobjective evolutionary algorithms on many-objective knapsack problems,'' \emph{{IEEE} Trans. Evol. Comput.}, vol.~19, no.~2, pp. 264--283, 2015.

\bibitem{visionMOP}
D.~J. Walker, R.~Everson, and J.~E. Fieldsend, ``Visualizing mutually nondominating solution sets in many-objective optimization,'' \emph{IEEE Transactions on Evolutionary Computation}, vol.~17, no.~2, pp. 165--184, 2012.

\bibitem{maoeasurvey}
K.~Li, R.~Wang, T.~Zhang, and H.~Ishibuchi, ``Evolutionary many-objective optimization: A comparative study of the state-of-the-art,'' \emph{IEEE Access}, vol.~6, pp. 26\,194--26\,214, 2018.

\bibitem{nsga3}
K.~Deb and H.~Jain, ``An evolutionary many-objective optimization algorithm using reference-point-based nondominated sorting approach, part {I:} solving problems with box constraints,'' \emph{{IEEE} Trans. Evol. Comput.}, vol.~18, no.~4, pp. 577--601, 2014.

\bibitem{nsga3survey}
C.~Gong, Y.~Nan, M.~Pang, H.~Ishibuchi, and Q.~Zhang, ``Performance of nsga-iii on multi-objective combinatorial optimization problems heavily depends on its implementations,'' in \emph{Proceedings of the Genetic and Evolutionary Computation Conference}, 2024, pp. 511--519.

\bibitem{gpgpu}
K.~S. Perumalla, ``Discrete-event execution alternatives on general purpose graphical processing units ({GPGPUs}),'' in \emph{20th Workshop on Principles of Advanced and Distributed Simulation (PADS'06)}.\hskip 1em plus 0.5em minus 0.4em\relax IEEE, 2006, pp. 74--81.

\bibitem{pytorch}
A.~Paszke, S.~Gross, F.~Massa, A.~Lerer, J.~Bradbury, G.~Chanan, T.~Killeen, Z.~Lin, N.~Gimelshein, L.~Antiga, A.~Desmaison, A.~Kopf, E.~Yang, Z.~DeVito, M.~Raison, A.~Tejani, S.~Chilamkurthy, B.~Steiner, L.~Fang, J.~Bai, and S.~Chintala, ``{PyTorch}: An imperative style, high-performance deep learning library,'' in \emph{Advances in Neural Information Processing Systems}, H.~Wallach, H.~Larochelle, A.~Beygelzimer, F.~d\textquotesingle Alch\'{e}-Buc, E.~Fox, and R.~Garnett, Eds., vol.~32.\hskip 1em plus 0.5em minus 0.4em\relax Curran Associates, Inc., 2019.

\bibitem{jax}
\BIBentryALTinterwordspacing
J.~Bradbury, R.~Frostig, P.~Hawkins, M.~J. Johnson, C.~Leary, D.~Maclaurin, G.~Necula, A.~Paszke, J.~VanderPlas, S.~Wanderman-Milne, and Q.~Zhang, ``{JAX}: Composable transformations of {P}ython+{N}um{P}y programs,'' 2018. [Online]. Available: \url{http://github.com/jax-ml/jax}
\BIBentrySTDinterwordspacing

\bibitem{tensorneat}
L.~Wang, M.~Zhao, E.~Liu, K.~Sun, and R.~Cheng, ``Tensorized neuroevolution of augmenting topologies for {GPU} acceleration,'' in \emph{Proceedings of the Genetic and Evolutionary Computation Conference, {GECCO} 2024, Melbourne, VIC, Australia, July 14-18, 2024}, X.~Li and J.~Handl, Eds.\hskip 1em plus 0.5em minus 0.4em\relax {ACM}, 2024.

\bibitem{tensorRVEA}
Z.~Liang, T.~Jiang, K.~Sun, and R.~Cheng, ``{GPU}-accelerated evolutionary multiobjective optimization using tensorized {RVEA},'' in \emph{Proceedings of the Genetic and Evolutionary Computation Conference, {GECCO} 2024, Melbourne, VIC, Australia, July 14-18, 2024}, X.~Li and J.~Handl, Eds.\hskip 1em plus 0.5em minus 0.4em\relax {ACM}, 2024.

\bibitem{vectorNSGA2}
A.~Aguilar{-}Rivera, ``A {GPU} fully vectorized approach to accelerate performance of {NSGA-2} based on stochastic non-domination sorting and grid-crowding,'' \emph{Appl. Soft Comput.}, vol.~88, p. 106047, 2020.

\bibitem{tensor}
S.~Lei, X.~Xiao, Y.~Gong, Y.~Li, and J.~Zhang, ``Tensorial evolutionary computation for spatial optimization problems,'' \emph{{IEEE} Trans. Artif. Intell.}, vol.~5, no.~1, pp. 154--166, 2024.

\bibitem{MaMOPSurvey}
B.~Li, J.~Li, K.~Tang, and X.~Yao, ``Many-objective evolutionary algorithms: {A} survey,'' \emph{{ACM} Comput. Surv.}, vol.~48, no.~1, pp. 13:1--13:35, 2015.

\bibitem{epsdomi}
M.~Laumanns, L.~Thiele, K.~Deb, and E.~Zitzler, ``Combining convergence and diversity in evolutionary multiobjective optimization,'' \emph{Evol. Comput.}, vol.~10, no.~3, pp. 263--282, 2002.

\bibitem{popstudy2}
T.~Chen, K.~Tang, G.~Chen, and X.~Yao, ``A large population size can be unhelpful in evolutionary algorithms,'' \emph{Theoretical Computer Science}, vol. 436, pp. 54--70, 2012.

\bibitem{positivepop3}
H.~E. Aguirre, A.~Liefooghe, S.~V{\'{e}}rel, and K.~Tanaka, ``A study on population size and selection lapse in many-objective optimization,'' in \emph{Proceedings of the {IEEE} Congress on Evolutionary Computation, {CEC} 2013, Cancun, Mexico, June 20-23, 2013}.\hskip 1em plus 0.5em minus 0.4em\relax {IEEE}, 2013, pp. 1507--1514.

\bibitem{earlyEA}
O.~Maitre, L.~A. Baumes, N.~Lachiche, A.~Corma, and P.~Collet, ``Coarse grain parallelization of evolutionary algorithms on {GPGPU} cards with easea,'' in \emph{Proceedings of the 11th Annual Conference on Genetic and Evolutionary Computation}, ser. GECCO '09.\hskip 1em plus 0.5em minus 0.4em\relax Association for Computing Machinery, 2009, p. 1403–1410.

\bibitem{gasrea}
D.~Sharma and P.~Collet, ``{GPGPU}-compatible archive based stochastic ranking evolutionary algorithm {(G-ASREA)} for multi-objective optimization,'' in \emph{Parallel Problem Solving from Nature - {PPSN} XI, 11th International Conference, Krak{\'{o}}w, Poland, September 11-15, 2010. Proceedings, Part {II}}, ser. Lecture Notes in Computer Science, R.~Schaefer, C.~Cotta, J.~Kolodziej, and G.~Rudolph, Eds., vol. 6239.\hskip 1em plus 0.5em minus 0.4em\relax Springer, 2010, pp. 111--120.

\bibitem{gpundsort}
S.~Gupta and G.~Tan, ``A scalable parallel implementation of evolutionary algorithms for multi-objective optimization on {GPUs},'' in \emph{{IEEE} Congress on Evolutionary Computation, {CEC} 2015, Sendai, Japan, May 25-28, 2015}.\hskip 1em plus 0.5em minus 0.4em\relax {IEEE}, 2015, pp. 1567--1574.

\bibitem{cudaalgos}
J.~J. Moreno, G.~Ortega, E.~Filatovas, J.~A. Mart{\'\i}nez, and E.~M. Garz{\'o}n, ``Using low-power platforms for evolutionary multi-objective optimization algorithms,'' \emph{The Journal of Supercomputing}, vol.~73, pp. 302--315, 2017.

\bibitem{masterslave}
M.~Yagoubi and M.~Schoenauer, ``Asynchronous master/slave moeas and heterogeneous evaluation costs,'' in \emph{Genetic and Evolutionary Computation Conference, {GECCO} '12, Philadelphia, PA, USA, July 7-11, 2012}, T.~Soule and J.~H. Moore, Eds.\hskip 1em plus 0.5em minus 0.4em\relax {ACM}, 2012, pp. 1007--1014.

\bibitem{mocell}
A.~J. Nebro, J.~J. Durillo, F.~Luna, B.~Dorronsoro, and E.~Alba, ``Mocell: {A} cellular genetic algorithm for multiobjective optimization,'' \emph{Int. J. Intell. Syst.}, vol.~24, no.~7, pp. 726--746, 2009.

\bibitem{GPUsurvey}
J.~R. Cheng and M.~Gen, ``Accelerating genetic algorithms with {GPU} computing: A selective overview,'' \emph{Computers \& Industrial Engineering}, vol. 128, pp. 514--525, 2019.

\bibitem{matrixEC}
Z.~Zhan, J.~Zhang, Y.~Lin, J.~Li, T.~Huang, X.~Guo, F.~Wei, S.~Kwong, X.~Zhang, and R.~You, ``Matrix-based evolutionary computation,'' \emph{{IEEE} Trans. Emerg. Top. Comput. Intell.}, vol.~6, no.~2, pp. 315--328, 2022.

\bibitem{evomo}
Z.~Liang, H.~Li, N.~Yu, K.~Sun, and R.~Cheng, ``Bridging evolutionary multiobjective optimization and {GPU} acceleration via tensorization,'' \emph{IEEE Transactions on Evolutionary Computation}, 2025.

\bibitem{rvea}
R.~Cheng, Y.~Jin, M.~Olhofer, and B.~Sendhoff, ``A reference vector guided evolutionary algorithm for many-objective optimization,'' \emph{{IEEE} Trans. Evol. Comput.}, vol.~20, no.~5, pp. 773--791, 2016.

\bibitem{evox}
B.~Huang, R.~Cheng, Z.~Li, Y.~Jin, and K.~C. Tan, ``{EvoX}: A distributed {GPU}-accelerated framework for scalable evolutionary computation,'' \emph{IEEE Transactions on Evolutionary Computation}, 2024.

\bibitem{dtlz}
K.~Deb, L.~Thiele, M.~Laumanns, and E.~Zitzler, ``Scalable multi-objective optimization test problems,'' in \emph{Proceedings of the 2002 Congress on Evolutionary Computation, {CEC} 2002, Honolulu, HI, USA, May 12-17, 2002}.\hskip 1em plus 0.5em minus 0.4em\relax {IEEE}, 2002, pp. 825--830.

\bibitem{molp}
H.~E. Aguirre and K.~Tanaka, ``Insights on properties of multiobjective mnk-landscapes,'' in \emph{Proceedings of the {IEEE} Congress on Evolutionary Computation, {CEC} 2004, 19-23 June 2004, Portland, OR, {USA}}.\hskip 1em plus 0.5em minus 0.4em\relax {IEEE}, 2004, pp. 196--203.

\bibitem{mokp}
C.~Bazgan, H.~Hugot, and D.~Vanderpooten, ``Solving efficiently the 0-1 multi-objective knapsack problem,'' \emph{Comput. Oper. Res.}, vol.~36, no.~1, pp. 260--279, 2009.

\bibitem{brax}
C.~D. Freeman, E.~Frey, A.~Raichuk, S.~Girgin, I.~Mordatch, and O.~Bachem, ``Brax - {A} differentiable physics engine for large scale rigid body simulation,'' 2021.

\bibitem{tase}
Q.~Wang, L.~Zhang, S.~Wei, and B.~Li, ``Tensor decomposition-based alternate sub-population evolution for large-scale many-objective optimization,'' \emph{Inf. Sci.}, vol. 569, pp. 376--399, 2021.

\bibitem{tfpso}
Q.~Wang, L.~Zhang, S.~Wei, B.~Li, and Y.~Xi, ``Tensor factorization-based particle swarm optimization for large-scale many-objective problems,'' \emph{Swarm Evol. Comput.}, vol.~69, p. 100995, 2022.

\bibitem{tensor-moead}
X.~Wang, Y.~Zhao, L.~Tang, and X.~Yao, ``{MOEA/D} with spatial-temporal topological tensor prediction for evolutionary dynamic multiobjective optimization,'' \emph{IEEE Transactions on Evolutionary Computation}, pp. 1--1, 2024.

\bibitem{platemo}
Y.~Tian, R.~Cheng, X.~Zhang, and Y.~Jin, ``Platemo: {A} {MATLAB} platform for evolutionary multi-objective optimization [educational forum],'' \emph{{IEEE} Comput. Intell. Mag.}, vol.~12, no.~4, pp. 73--87, 2017.

\bibitem{pymoo}
J.~Blank and K.~Deb, ``Pymoo: Multi-objective optimization in python,'' \emph{IEEE Access}, vol.~8, pp. 89\,497--89\,509, 2020.

\end{thebibliography}
\clearpage  
\onecolumn 
\appendix
\subsection{Tensor-based Evolutionary Algorithms}
Recent advancements in tensor-based evolutionary computation (EC) have demonstrated significant potential for enhancing computational efficiency and scalability. 
Notable developments include several tensor-based methods. TASE~\cite{tase} applied tensor decomposition to implement alternating subpopulation evolution. TFPSO~\cite{tfpso} extended particle swarm optimization with tensor models for multiobjective and dynamic optimization problems. STT-DMOEA/D~\cite{tensor-moead} introduced a spatiotemporal topological tensor model for adaptive population initialization. Klosko et al.\cite{tensor} and Zhan et al.\cite{matrixEC} independently introduced fundamental tensor computation models for EC in 2022. Most recently, TensorRVEA~\cite{tensorRVEA} introduced the first fully tensorized implementation of a reference vector-guided MOEA based on RVEA~\cite{rvea}.
Tensor-based methods align well with GPU hardware, reducing architectural complexity and enabling efficient parallelism. However, most existing methods incorporate tensors without fully exploiting GPU acceleration. GPU-accelerated tensor methods remain in their infancy.

\subsection{NSGA-III Implementation}
There is no standard implementation of NSGA-III since no source code has been provided by the authors of the NSGA-III paper. Notable implementations include PlatEMO and pymoo. Our comparisons include several NSGA-III implementations on CPU and GPU, referencing both PlatEMO~\cite{platemo} and pymoo~\cite{pymoo} (a Python-based framework). However, since pymoo could not support more than 400 individuals in our experiments (leading to memory issues at larger scales), it is excluded from our experiments.

Specifically, we implemented and compared:
\begin{itemize}
    \item \textbf{NSGA-III on CPU}: The original NSGA-III referencing the PlatEMO platform running on CPU.
    \item \textbf{NSGA-III on GPU}: A direct GPU port of the above CPU-based code, with minimal modifications and optimized by XLA, a compiler for GPU acceleration.
    \item \textbf{TensorNSGA-III on GPU}: Our tensorized implementation version of NSGA-III
    \item \textbf{TensorRVEA on GPU}: Another tensorized algorithm approximating RVEA, using reference points but eliminating the while-loop.
\end{itemize}

\subsection{Cache Mechanism Analysis}
TensorNSGA-III's cache tensor mechanism introduces additional computational operations compared to NSGA-III (CPU) through full-population candidate evaluation in niche selection. However, massive GPU parallelization more than compensates for this increased workload, resulting in net runtime reductions.
The NSGA-III on GPU implementation suffers from sequential bottlenecks during frequent selection operations, particularly in niche selection, where the selection process degenerates into serial computations. Given the simpler architecture and lower clock frequencies of GPU cores compared to CPUs, this minimally modified GPU version underperforms even the CPU version in practice. 
The cache tensor strategy embodies a classic space-time tradeoff, where increased memory consumption enables computational acceleration. In our experiments, GPU memory constraints did not significantly impact performance at tested population scales.

\subsection{Acceleration Performance on Varying Objectives}
To validate the acceleration performance of proposed TensorNSGA-III when handling different objectives on large-scale MaOPs, we compare NSGA-III on CPU, NSGA-III on GPU, TensorNSGA-III on GPU and TensorRVEA on GPU. For these experiments, we used the DTLZ3 problem with population size 800 and 1000 decision variables. The objectives from 4 to 512, and each algorithm was run for 100 generations.

Table~\ref{tab:appd_dtlz2} summarizes the average runtime per generation, highlighting TensorNSGA-III's superior scalability across objective dimensions. Its tensorized implementation maintains runtimes within milliseconds even with 128 objectives, achieving up to $229\times$ speedup over the CPU baseline. In contrast, the GPU version of the original NSGA-III performs worse than the CPU version and shows the highest instability at 32 objectives.

\begin{table*}[h!]
    \centering
    \small
    \caption{Comparison of average runtime per generation of algorithms on DTLZ3 problem with varying objectives. Speedup values are compared to NSGA-III on CPU.}
    \label{tab:appd_dtlz2}
    \setlength{\tabcolsep}{5pt}
    \renewcommand{\arraystretch}{1.2}
    \begin{tabular}{l c c c c}
    \hline
    \textbf{Objectives} & \textbf{NSGA-III on GPU} & \textbf{NSGA-III on CPU} & \textbf{TensorNSGA-III} & \textbf{Speedup} \\ \hline
    4 & $\scientific{1.295}{+1} \pm \scientific{6.267}{-3}$ & $\scientific{9.219}{-1} \pm \scientific{2.349}{-3}$ & \cellcolor{gray!20}{$\scientific{4.334}{-3} \pm \scientific{3.713}{-5}$} & $213$ \\ \hline
    8 & $\scientific{1.522}{+1} \pm \scientific{8.392}{-3}$ & $\scientific{1.061}{+0} \pm \scientific{2.741}{-3}$ & \cellcolor{gray!20}{$\scientific{4.624}{-3} \pm \scientific{2.922}{-5}$} & $229$ \\ \hline
    16 & $\scientific{6.366}{+0} \pm \scientific{3.128}{-3}$ & $\scientific{4.722}{-1} \pm \scientific{1.179}{-3}$ & \cellcolor{gray!20}{$\scientific{5.083}{-3} \pm \scientific{3.836}{-5}$} & $93$\\ \hline
    32 & $\scientific{1.317}{+1} \pm \scientific{4.760}{-3}$ & $\scientific{8.684}{-1} \pm \scientific{2.147}{-3}$ & \cellcolor{gray!20}{$\scientific{6.514}{-3} \pm \scientific{1.346}{-4}$} & $133$ \\ \hline
    64 & $\scientific{3.050}{+0} \pm \scientific{1.555}{-3}$ & $\scientific{4.784}{-1} \pm \scientific{2.498}{-3}$ & \cellcolor{gray!20}{$\scientific{8.103}{-3} \pm \scientific{6.638}{-5}$} & $59$  \\ \hline
    128 & $\scientific{6.040}{+0} \pm \scientific{2.330}{-3}$ & $\scientific{9.282}{-1} \pm \scientific{7.251}{-3}$ & \cellcolor{gray!20}{$\scientific{1.016}{-2} \pm \scientific{6.324}{-5}$} & $91$\\ \hline
    256 & $\scientific{1.215}{+1} \pm \scientific{4.058}{-3}$ & $\scientific{2.012}{+0} \pm \scientific{8.756}{-3}$ & \cellcolor{gray!20}{$\scientific{2.348}{-2} \pm \scientific{8.348}{-4}$} & $86$ \\ \hline
    512 & $\scientific{1.239}{+1} \pm \scientific{4.675}{-3}$ & $\scientific{3.918}{+0} \pm \scientific{4.190}{-2}$ & \cellcolor{gray!20}{$\scientific{3.199}{-2} \pm \scientific{7.809}{-4}$} & $123$  \\ \hline
    \end{tabular}
\end{table*}

\subsection{Hypervolume Calculation}
For discrete MaOPs, the HV metric is influenced by both the reference point and the ideal point. These points are calculated as follows:

\begin{equation}
\label{eq:ref_point}
ref_j = 1.01 \times f_j^{\max}, \quad j=1,2,\dots, m,
\end{equation}

\begin{equation}
\label{eq:ideal_point}
ideal_j = 0.9 \times f_j^{\min}, \quad j=1,2,\dots, m,
\end{equation}

\begin{equation}
\label{eq:hv_max}
HV_{max} = \prod_{j=1}^{m} (ref_j - ideal_j).
\end{equation}

Here, \( f_j^{\max} \) and \( f_j^{\min} \) represent the maximum and minimum values of the \( j \)-th objective across all Pareto Fronts (PFs) considered in our experiments, respectively. The final HV value for each PF is obtained by dividing its calculated HV by \( HV_{max} \) to normalize the metric:

\begin{equation}
\label{eq:normalized_hv}
HV_{normalized} = \frac{HV}{HV_{max}}.
\end{equation}

This normalization ensures that the HV values are scaled between 0 and 1, facilitating consistent comparisons across different problem instances and objective spaces.

\subsection{Brax}
Brax is a physics simulator widely used for reinforcement learning but generally focuses on single-objective tasks. According to TensorRVEA, we decompose single-objective reward signals (e.g., Swimmer, Halfcheetah, and Hopper) into multiple sub-rewards, thus forming multiobjective control problems.

Table~\ref{tab:brax} summarizes the chosen tasks. Each environment was converted to two-objective or three-objective forms, depending on how we split the individual reward components. And all problems are translated into minimum optimization problems.

\begin{table}[hb]
  \caption{Summary of multiobjective robot control tasks}
  \label{tab:brax}
  \centering
  \renewcommand{\arraystretch}{1.2}  
  \begin{tabular}{l c c p{5.5cm}}
    \toprule 
    Problem & Objectives & Dimensions & Description of Objectives \\ \hline
    \midrule 
    MoSwimmer & 2 & 178 & Forward Reward, Control Cost \\ \hline
    MoHalfcheetah & 2 & 390 & Forward Reward, Control Cost \\ \hline
    MoReacher & 2 & 226 & Distance Reward, Control Cost \\ \hline
    \bottomrule
  \end{tabular}
\end{table}

\end{document}